\documentclass[letterpaper, 10 pt, journal, twoside]{ieeetran}
\IEEEoverridecommandlockouts    %
\usepackage[utf8]{inputenc} % added by arXiv
\usepackage{multicol}
\usepackage[bookmarks=true]{hyperref} %

\newtheorem{theorem}{Theorem}

\newtheorem{assumption}{Assumption}
\newtheorem{remark}{Remark}

\usepackage{xcolor}
\usepackage{amsmath}
\usepackage{bbm}
\usepackage{graphicx}
\usepackage{amssymb}
\usepackage[detect-weight=true, detect-family=true]{siunitx}
\usepackage[normalem]{ulem}

\DeclareSIUnit{\million}{M}
\DeclareSIUnit{\thousand}{k}

\usepackage{float}
\usepackage{censor}
\usepackage{nicefrac}

\usepackage[natbib=true,
    style=ieee,
    citestyle=numeric-comp,
    backend=biber,
    maxbibnames=5,
    maxcitenames=99,
    doi=false,
    url=false,
    giveninits=true]{biblatex}
\DeclareSourcemap{
  \maps[datatype=bibtex]{
    \map{
      \step[fieldsource=url,
        match=\regexp{\\_},
        replace=\regexp{_}]
    }
  }
}

\AtBeginBibliography{\footnotesize}
\newbibmacro{string+url}[1]{%
  \iffieldundef{url}{%
  \iffieldundef{doi}{#1}{\href{http://dx.doi.org/\thefield{doi}}{#1}}}{\href{\thefield{url}}{#1}}}
\DeclareFieldFormat*{title}{\usebibmacro{string+url}{``#1''}}

\usepackage{tikz}
\usepackage{textcomp}
\usepackage{hyperref}
\usepackage{lipsum}
\newcommand\copyrighttext{%
  \footnotesize \textcopyright 2020 IEEE. Personal use of this material is permitted.
  Permission from IEEE must be obtained for all other uses, in any current or future 
  media, including reprinting/republishing this material for advertising or promotional 
  purposes, creating new collective works, for resale or redistribution to servers or 
  lists, or reuse of any copyrighted component of this work in other works. 
  DOI: see top of this page.}
\newcommand\copyrightnotice{%
\begin{tikzpicture}[remember picture,overlay]
\node[anchor=south,yshift=7.5pt] at (current page.south) {\fbox{\parbox{\dimexpr\textwidth-\fboxsep-\fboxrule\relax}{\copyrighttext}}};
\end{tikzpicture}%
}

\usepackage[labelformat=simple]{subcaption}
\DeclareCaptionLabelSeparator{periodspace}{.\quad}
\newcommand{\argmin}{\text{argmin}}
\newcommand{\vs}{\mathbf{s}}
\newcommand{\vp}{\mathbf{p}}
\newcommand{\vg}{\mathbf{g}}
\newcommand{\vv}{\mathbf{v}}
\newcommand{\vu}{\mathbf{u}}
\newcommand{\ve}{\mathbf{e}}
\newcommand{\vo}{\mathbf{o}}
\newcommand{\vx}{\mathbf{x}}
\newcommand{\vk}{\mathbf{k}}

\newcommand{\cG}{\mathcal{G}}
\newcommand{\cV}{\mathcal{V}}
\newcommand{\cS}{\mathcal{S}}
\newcommand{\cSO}{\mathcal{S}_0}
\newcommand{\cOmega}{\mathit{\Omega}}
\newcommand{\cU}{\mathcal{U}}
\newcommand{\cO}{\mathcal{O}}
\newcommand{\cX}{\mathcal{X}}
\newcommand{\cN}{\mathcal{N}}
\newcommand{\cI}{\mathcal{I}}
\newcommand{\cD}{\mathcal{D}}

\newcommand{\fu}{\mathit{u}}
\newcommand{\fy}{\mathit{y}}
\newcommand{\fc}{\mathit{c}}
\newcommand{\ff}{\mathit{f}}
\newcommand{\fapi}{\mathit{\alpha_\pi}}
\newcommand{\fb}{\mathit{b}}
\newcommand{\fpi}{\mathit{\pi}}
\newcommand{\fphi}{\mathit{\phi}}
\newcommand{\frho}{\mathit{\rho}}
\newcommand{\fPsi}{\mathit{\Psi}}
\newcommand{\fsigma}{\mathit{\sigma}}
\newcommand{\fFF}{\mathit{FF}}
\newcommand{\fg}{\mathit{g}}
\newcommand{\fh}{\mathit{h}}
\newcommand{\fpsi}{\mathit{\psi}}

\newcommand{\ji}{^{ij}}
\renewcommand{\th}{^{\text{th}}}

\begin{document}

\title{
GLAS: Global-to-Local Safe Autonomy Synthesis for Multi-Robot Motion Planning with End-to-End Learning
}

\author{Benjamin Rivière, Wolfgang  H\"onig, Yisong Yue, and Soon-Jo Chung
\thanks{Manuscript received: February, 24, 2020; Accepted April, 20, 2020.} %
\thanks{The authors are with California Institute of Technology, USA.
\texttt{\{briviere, whoenig, yyue, sjchung\}@caltech.edu}.}
\thanks{This paper was recommended for publication by Editor Nak Young Chong upon evaluation of the Associate Editor and Reviewers' comments.
This work was supported by the Raytheon Company and Caltech/NASA Jet Propulsion Laboratory. Video: \url{https://youtu.be/z9LjSfLfG6c}. Code: \url{https://github.com/bpriviere/glas}.} %
\thanks{Digital Object Identifier (DOI): see top of this page.}
}

\maketitle

\copyrightnotice

\begin{abstract}
We present GLAS: \underline{G}lobal-to-\underline{L}ocal \underline{A}utonomy \underline{S}ynthesis, a provably-safe, automated distributed policy generation for multi-robot motion planning. Our approach combines the advantage of centralized planning of avoiding local minima with the advantage of decentralized controllers of scalability and distributed computation. In particular, our synthesized policies only require relative state information of nearby neighbors and obstacles, and compute a provably-safe action.
Our approach has three major components: i) we generate demonstration trajectories using a global planner and extract local observations from them, ii) we use deep imitation learning to learn a decentralized policy that can run efficiently online, and iii) we introduce a novel differentiable safety module to ensure collision-free operation, thereby allowing for end-to-end policy training. Our numerical experiments demonstrate that our policies have a \SI{20}{\percent} higher success rate than optimal reciprocal collision avoidance, ORCA, across a wide range of robot and obstacle densities. We demonstrate our method on an aerial swarm, executing the policy on low-end microcontrollers in real-time. 
\end{abstract}


\IEEEpeerreviewmaketitle

\section{Introduction}

\IEEEPARstart{T}{eams} of robots that are capable of navigating in dynamic and occluded environments are important for applications in next generation factories, urban search and rescue, and formation flying in cluttered environments or in space.
Current centralized approaches can plan such motions with completeness guarantees, but require full state information not available to robots on-board, and are too computationally expensive to run in real-time.
Distributed approaches instead use local decoupled optimization, but  often cause robots to get trapped in local minima in cluttered environments.
Our approach, GLAS, bridges this gap by using a global planner offline to learn a decentralized policy that can run efficiently online.
We can thus automatically synthesize an efficient policy that avoids getting trapped in many cases.
Unlike other learning-based methods for motion planning, GLAS operates in continuous state space with a time-varying number of neighbors and generates provably safe, dynamically-coupled policies.  
We demonstrate in simulation that our policy achieves significantly higher success rates compared to ORCA, a state-of-the-art decentralized approach for single integrator dynamics.
We also extend our approach to double integrator dynamics, and demonstrate that our synthesized policies work well on a team of quadrotors with low-end microcontrollers.

\begin{figure}[t]
	\centering
    \includegraphics[width=0.95\linewidth]{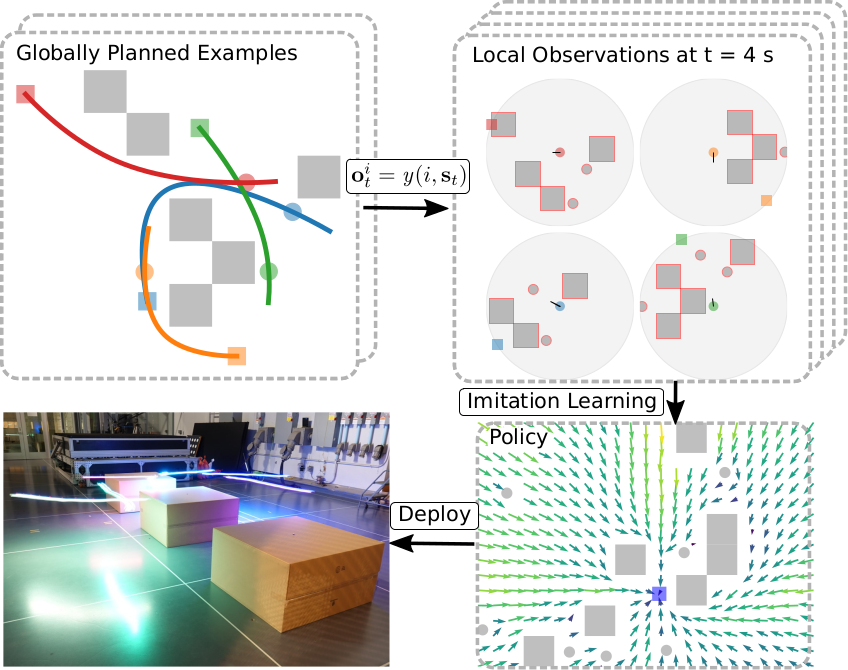}
    \caption{We learn distributed policies using trajectories from a global planner. We mask non-local information with an observation model, and perform imitation learning on an observation-action dataset. In the policy, the grey boxes are static obstacles, the grey circles are other robots, and the blue square is the robot's goal position.}
    \label{fig:overview}
\end{figure}

The overview of GLAS is shown in Fig.~\ref{fig:overview}. First, we generate trajectories for random multi-robot motion planning instances using a global planner.  Second, we apply a local observation model to generate a dataset of observation-action pairs. Third, we perform deep imitation learning to generate a local policy that imitates the expert (the global policy) to avoid local minima. The vector field of Fig.~\ref{fig:overview} shows that the policy successfully avoids local minima traps between obstacles and gridlock between robots.

We guarantee the safety of our policy through a convex combination of the learned desired action and our safety module. Instead of using existing optimization-based methods, we derive an analytic form for our safety module. The advantage of this safety module is that it is fully differentiable, which enables us to train our policy in an end-to-end fashion, resulting in policies that use less control effort.

Our main contributions are: i) to our knowledge, this is the first approach that automatically synthesizes a local policy from global demonstrations in a continuous state/action domain while guaranteeing safety and ii) derivation of a novel differentiable safety module compatible with end-to-end learning for dynamically-coupled motion planning.

We show that our  policies outperform the state-of-the-art distributed multi-robot motion planners for single integrator dynamics in terms of success rate in a wide range of robot/obstacle density cases. We also implement GLAS for more physically-realistic double integrator dynamics in both simulation and experimentation with an aerial swarm, demonstrating real-time computation on low-end microcontrollers.

\subsection{Related Work}
Multi-robot motion planning is an active area of research because it is a non-convex optimization problem with high state and action dimensionality. We compare the present work with state-of-the-art methods: (a) collision avoidance controllers, (b) optimal motion-planners, and (c) deep-learning methods.

\textit{Collision Avoidance:} Traditional controller-level approaches include Optimal Reciprocal Collision Avoidance (ORCA)~\cite{ORCA}, Buffered Voronoi Cells~\cite{bandyopadhyay2017probabilistic}, Artificial Potential Functions~\cite{khatib_1990,Rimon_1992, Tanner_2005}, and Control Barrier Functions~\cite{WangAE17}. These methods are susceptible to trapping robots in local minima. We address this problem explicitly by imitating a complete global planner with local information. For optimal performance, we propose to learn a controller end-to-end, including the safety module. Existing methods that are based on optimization~\cite{ORCA,bandyopadhyay2017probabilistic,WangAE17} are challenging for backpropagation for end-to-end training. Existing analytic methods~\cite{khatib_1990} do not explicitly consider gridlocks, where robots' respective barriers cancel each other and disturbances can cause the system to violate safety. Thus, we derive a novel differentiable safety module. This system design of fully differentiable modules for end-to-end backpropagation is also explored in reinforcement learning~\cite{cheng_2019b} and estimation~\cite{Jonschkowski_18}. 

\textit{Motion Planners:} Motion planners are a higher-level approach that explicitly solves the optimal control problem over a time horizon. Solving the optimal control problem is non-convex, so most recent works with local guarantees use approximate methods like Sequential Convex Programming to quickly reach a solution~\cite{SATO,Hoenig_2018}. Motion planners are distinguished as either global and centralized~\cite{Hoenig_2018} or local and decentralized~\cite{LuisS19,SATO}, depending on whether they find solutions in joint space or computed by each robot. 

GLAS is inherently scalable because it is computed at each robot. Although it has no completeness guarantee, we empirically show it avoids local minima more often than conventional local methods. The local minima issue shared by all the local methods is a natural trade-off of decentralized algorithms. Our method explores this trade-off explicitly by imitating a complete, global planner with only local information. 

\textit{Deep Learning Methods:} Recently, there have been new learning-based approaches for  multi-robot path planning~\cite{Sartoretti_2019,Li_2019,khan_2019a,khan_2019b,raju_2019}. These works use deep Convolutional Neural Networks (CNN) in a discrete state/action domain. Such discretization prevents coupling to higher-order robot dynamics whereas our solution permits tight coupling to the system dynamics by operating in a continuous state/action domain using a novel network architecture based on Deep Sets~\cite{deepsets_2017}. Deep Sets are a relatively compact representation that leverages the permutation-invariant nature of the underlying interactions, resulting in a less computationally-expensive solution than CNNs. In contrast to our work, the Neural-Swarm approach~\cite{neural-swarm} uses Deep Sets to augment a tracking controller for close proximity flight.

Imitation Learning (IL) can imitate an expensive planner~\cite{pan_18,Li_2019,khan_2019a,Sartoretti_2019}, thereby replacing the optimal control or planning solver with a function that approximates the solution. GLAS uses IL and additionally changes the input domain from full state information to a local observation, thereby enabling us to synthesize a decentralized policy from a global planner. 

\section{Problem Formulation}
\label{sec:problem_formulation}
\textit{Notation:} We denote vectors with a boldface lowercase, functions with an italics lowercase letter, scalar parameters with plain lowercase, subspaces/sets with calligraphic uppercase, and matrices with plain uppercase. We denote a robot index with a superscript $i$ or $j$, and a state or action without the robot index denotes a joint space variable of stacked robot states. A double superscript index denotes a relative vector, for example $\vs\ji = \vs^j-\vs^i$. We use a continuous time domain, and we suppress the explicit time dependency for states, actions, and dependent functions for notation simplicity. 

\textit{Problem Statement:} Let $\cV$ denote the set of $n_i$ robots, $\cG = \{\vg^1,\ldots,\vg^{n_i}\}$ denote their respective goal states, $\cSO = \{\vs^1_0,\ldots,\vs^{n_i}_0\}$ denote their respective start states, and $\cOmega$ denote the set of $m$ static obstacles. At time $t$, each robot $i$ makes a local observation, $\vo^i$, uses it to formulate an action, $\vu^i$, and updates its state, $\vs^i$, according to the dynamical model. Our goal is to find a controller, $\fu: \cO \rightarrow \cU$ that synthesizes actions from local observations through: 
\begin{align}
	\vo^i = \fy(i,\vs), \  \ \ \
	\vu^i = \fu(\vo^i), \ \ \ \forall i,t  
\end{align}
to approximate the solution to the optimal control problem: 
\begin{equation}
\begin{alignedat}{2}
	\vu^{*} &= \argmin_{\{\vu^i | \forall i,t\}} \fc(\vs,\vu) \quad 
	&\textrm{ s.t.} \\
		\dot{\vs}^i &= \ff(\vs^i,\vu^i) &\forall i,t \\ 
		\vs^i(0) &= \vs^i_0, \ \vs^i(t_f) = \vg^i, \ \vs \in \cX \ \ &\forall i \\ 
		\|\vu^i\|_2 & \leq u_\text{max} &\forall i,t
\end{alignedat}
\end{equation}
where $\cO$, $\cU$, and $\cS$ are the observation, action, and state space, $\fy$ is the local observation model, $\fc$ is some cost function, $\ff$ is the dynamical model, $\cX \subset \cS$ is the safe set capturing safety of all robots, $t_f$ is the time of the simulation, and $u_\text{max}$ is the maximum control bound. 

\textit{Dynamical Model:}
We consider both single and double integrator systems. The single integrator state and action are position and velocity vectors in $\mathbb{R}^{n_q}$, respectively.
The double integrator state is a stacked position and velocity vector in $\mathbb{R}^{2 n_q}$, and the  action is a vector of accelerations in $\mathbb{R}^{n_q}$. Here, $n_q$ is the dimension of the workspace. The dynamics of the $i^{\text{th}}$ robot for single and double integrator systems are:
\begin{alignat}{2}
	\dot{\vs}^i = \dot{\vp}^i = \vu^i 
	\quad\text{and} \quad
	\dot{\vs}^i = \begin{bmatrix} \dot{\vp}^i \\ \dot{\vv}^i \end{bmatrix} = \begin{bmatrix} \vv^i \\ \vu^i \end{bmatrix}, \label{eq:dynamics}
\end{alignat}
respectively, where $\vp^i$ and $\vv^i$ denote position and velocity. 

\textit{Observation Model:}
We are primarily focused on studying the transition from global to local, which is defined via an observation model, $\fy: \cV \times \cS \rightarrow \cO$. An observation is:
\begin{align}
	\vo^i &= \left[ \ve^{ii}, \{\vs\ji\}_{j \in \cN_{\cV}^i}, \{ \vs\ji \}_{j \in \cN_\cOmega^i} \right], \label{eq:observation}
\end{align}
where $\ve^{ii} = \vg^i - \vs^i$ and $\cN_{\cV}^i$, $\cN_{\cOmega}^i$ denote the neighboring set of robots and obstacles, respectively. These sets are defined by the observation radius, $r_\text{sense}$, e.g.,  
\begin{equation}
	\cN_\cV^i = \{ j \in \cV \ | \ \| \vp\ji \|_2 \leq r_\text{sense} \}.
\end{equation}
We encode two different neighbor sets because we input robots and obstacles through respective sub-networks of our neural network architecture in order to generate heterogeneous behavior in reaction to different neighbor types.
We denote the union of the neighboring sets as $\cN^i$.

\textit{Performance Metrics:}
To evaluate performance, we have two criteria as specified by the optimal control problem. We define our metrics over the set of successful robots, $\cI$, that reach their goal and have no collisions:
\begin{align}
    \cI &= \{ i \in \cV \ | \ \vs^i(t_f) = \vg^i \text{ and } \| \vp_t\ji \| > r_\text{safe}, \ \forall j,t \}.
\end{align}
Our first metric of success, $r_s$, is the number of successful robots, and our second metric, $r_p$, is the cost of deploying a successful robot trajectory. For example, if the cost function $\fc(\vs,\vu)$ is the total control effort, the performance metrics are: 
\begin{alignat}{2}
    r_s &= |\cI|, \ \ \ \mbox{and}\ \ \  r_p &= \sum_{i\in\cI} \int_0^{t_f} \|\vu^i\|_2 dt.
    \label{eq:metrics} 
\end{alignat}

\section{Algorithm Description and Analysis: GLAS}
\begin{figure}
	\includegraphics[width=\linewidth]{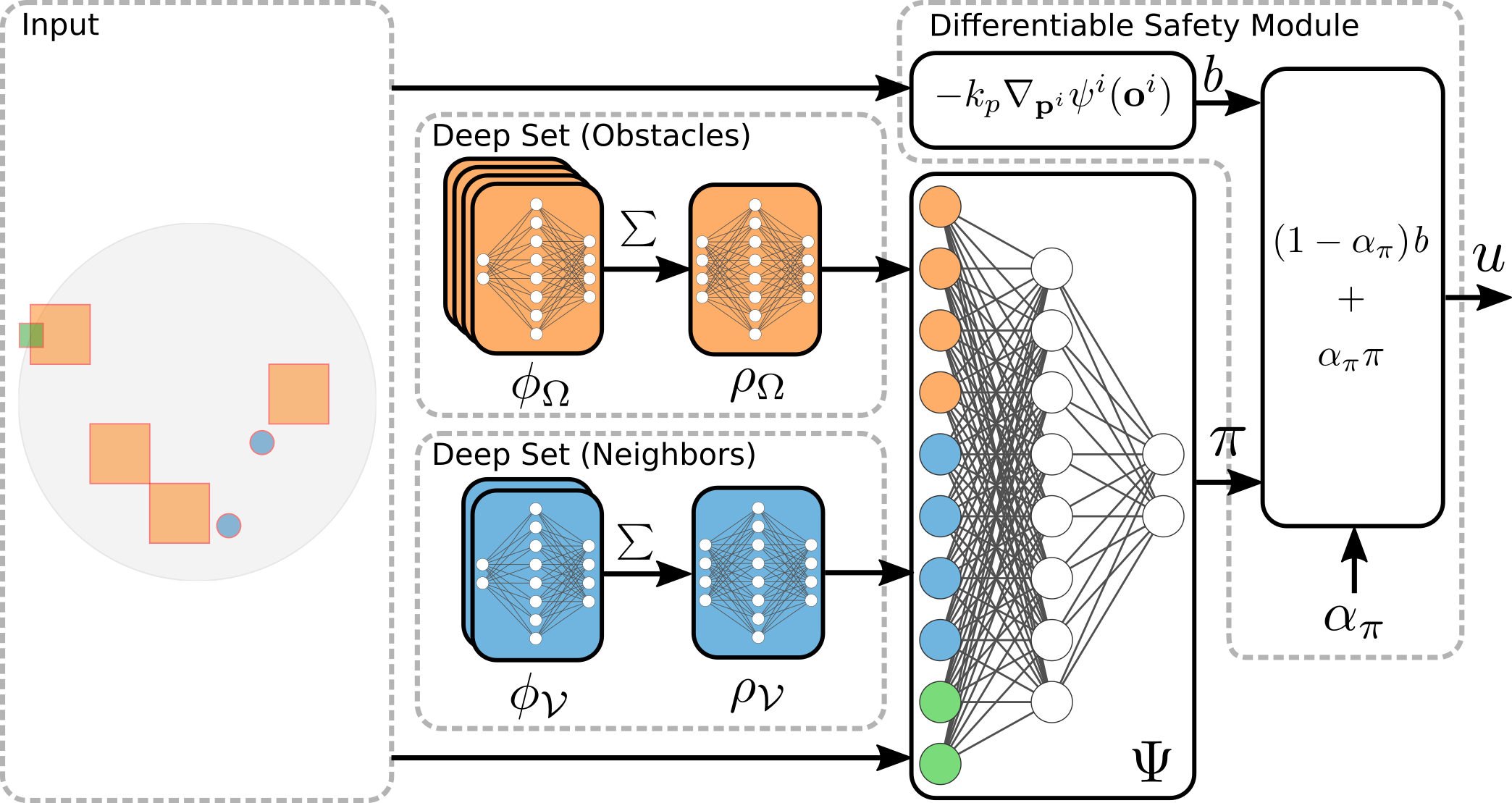}
	\caption{Neural network architecture consisting of 5 feed-forward components.  Each relative location of an obstacle is evaluated in $\fphi_\cOmega$; the sum of all $\fphi_\cOmega$ outputs is the input of $\frho_\cOmega$ (deep set for obstacles). Another deep set ($\fphi_\cV$ and $\frho_\cV$) is used for neighboring robot positions. The desired control $\fpi$ is computed by $\fPsi$, which takes the output of $\frho_\cOmega$, $\frho_\cV$, and the relative goal location as input. The actual number of hidden layers is larger than shown. The formula for the single integrator $\fb$ function is shown.}
	\label{fig:architecture}
\end{figure}

In this section, we derive the method of \textbf{GLAS}, \textbf{G}lobal-to-\textbf{L}ocal \textbf{A}utonomy \textbf{S}ynthesis, to find policy $\fu$: 
\begin{alignat}{2}
	\fu(\vo^i) &= \fapi(\vo^i) \fpi(\vo^i) + (1 - \fapi(\vo^i)) \fb(\vo^i), \label{eq:controller} 
\end{alignat}
where $\fpi: \cO \rightarrow \cU$ is a learned function, $\fb: \cO \rightarrow \cU$ is a safety control module to ensure safety, and $\fapi: \cO \rightarrow [0,1]$ is an adaptive gain function. We discuss each of the components of the controller in this section. The overview of our controller architecture is shown in Fig.~\ref{fig:architecture}.

\subsection{Neural Policy Synthesis via Deep Imitation Learning}
We describe how to synthesize the neural policy $\fpi$ that imitates the behavior of an expert, where the expert refers to a global optimal planner. Explicitly, we take batches of observation-action pairs from an expert demonstration dataset and we train a neural network by minimizing the loss on output action given an observation. To use this method, we need to generate an observation-action pair dataset from expert demonstration and design a deep learning architecture compatible with dynamic sensing network topologies. 

\subsubsection{Generating Demonstration Data}
Our dataset is generated using expert demonstrations from an existing centralized planner~\cite{Hoenig_2018}. This planner is resolution-complete and avoids local minima; it is computationally efficient so we can generate large expert demonstration datasets; and it uses an optimization framework that can minimize control effort, so the policy imitates a solution with high performance according to the previously defined metrics. Specifically, we create our dataset by generating maps with (i) fixed-size static obstacles with random uniformly sampled grid positions and (ii) start/goal positions for a variable number of robots, and then by computing trajectories using the centralized planner. For each timestep and robot, we retrieve the local observation, $\vo^i$ by masking the non-local information with the observation model $\fy$, and retrieving the action, $\vu^i$ through the appropriate derivative of the robot $i$ trajectory, see Fig.~\ref{fig:overview}. We repeat this process $n_\mathrm{case}$ times for each robot/obstacle case. Our dataset, $\cD$, is:
\begin{align}
	\cD &= \{ (\vo^i,\vu^i)_k \ | \ \forall i \in \cV, \forall k \in \{1\ldots n_\mathrm{case}\}, \forall t \}.
\end{align}

\subsubsection{Model Architecture with Deep Sets}
The number of visible neighboring robots and obstacles can vary dramatically during each operation, which causes the dimensionality of the observation vector to be time-varying. Leveraging the permutation invariance of the observation, we model variable number of robots and obstacles with the Deep Set architecture~\cite{deepsets_2017,neural-swarm}. Theorem 7 from~\cite{deepsets_2017} establishes this property:
\begin{theorem}
	Let $\ff: [0,1]^l \rightarrow \mathbb{R}$ be a permutation invariant continuous function iff it has the representation:
	\begin{align}
		f(x_1,\ldots,x_l) = \frho \left(\sum_{m=1}^l \fphi(x_m) \right),
	\end{align} 
	for some continuous outer and inner function $\frho:\mathbb{R}^{l+1}\rightarrow \mathbb{R}$ and $\fphi:\mathbb{R} \rightarrow \mathbb{R}^{l+1}$, respectively. The inner function $\fphi$ is independent of the function $\ff$. 
\end{theorem} 

Intuitively, the $\fphi$ function acts as a contribution from each element in the set, and the $\frho$ function acts to combine the contributions of each element. In effect, the policy can learn the contribution of the neighboring set of robots and obstacles with the following network structure: 
\begin{equation}
\begin{aligned}
	\fpi(\vo^i)_\mathrm{n} &= \fPsi( 
	[\frho_\cOmega(\!\!\sum_{j\in\cN_\cOmega^i} \fphi_\cOmega(\vs\ji)); \frho_\cV(\!\!\sum_{j\in\cN_V^i} \phi_\cV(\vs\ji))]), \\ 
	\fpi(\vo^i) &= \fpi(\vo^i)_\mathrm{n} \ \min\{ \nicefrac{\pi_\mathrm{max}}{\|\fpi(\vo^i)_\mathrm{n}\|_2}, 1\},
\end{aligned}
\end{equation}
where the semicolon denotes a stacked vector and $\fPsi,\frho_\cOmega,\fphi_\cOmega,\frho_\cV,\phi_\cV$ are feed-forward networks of the form:
\begin{align}
    \fFF(\vx) = W^{l}\fsigma(\ldots W^1 \fsigma(\vx)),
\end{align}
where $\fFF$ is a feed-forward network on input $\vx$, $W^l$ is the weight matrix of the $l\th$ layer, and $\fsigma$ is the  activation function. We define the parameters for each of the $5$ networks in Sec.~\ref{sec:experiments}. We also scale the output of the $\fpi$ network to always be less than $\pi_\text{max}$, to maintain consistency with our baselines. 

\subsubsection{End-to-End Training}
We train the neural policy $\fpi$ with knowledge of the safety module $\fb$, to synthesize a controller $\fu$ with symbiotic components. We train through the output of $\fu$, not $\fpi$, even though $\fb$ has no tunable parameters. In effect, the parameters of $\fpi$ are updated such that the policy $\fpi$ smoothly interacts with $\fb$ while imitating the global planner. With respect to a solution that trains through the output of $\fpi$, end-to-end learning generates solutions with lower control effort, measured through the $r_p$ metric~\eqref{eq:metrics}, see Fig.~\ref{fig:ss_ex}.

\subsubsection{Additional Methods in Training}
We apply additional preprocessing methods to the observation to improve our training process performance and to regularize the data. We denote the difference between the original observation and the preprocessed data with an apostrophe; e.g., the input to the neural network is an observation vector denoted by ${\vo^i}'$. 

We scale the relative goal vector observation as follows:
\begin{alignat}{3}
	{\ve^{ii}}' &= \alpha_g \ve^{ii}, \text{ where }
	\alpha_g &= \min \{ \nicefrac{r_\text{sense}}{\|\ve^{ii}\|} , 1 \}. 
\end{alignat}
This regularizes cases when the goal is beyond the  sensing radius. In such cases, the robot needs to avoid any robots/obstacles and continue toward the goal. However, the magnitude of $\ve^{ii}$ outside the sensing region is not important. 

We cap the maximum cardinality of the neighbor and obstacle sets with $\overline{\cN_\cV}$ and $\overline{\cN_\cOmega}$, e.g.,
\begin{equation}
	{\cN_\cV^i}' = \{ j \in \cN_\cV^i \ | \ \text{ $\overline{\cN_\cV}$-closest robots w.r.t. } \| \vp\ji \| \}.
\end{equation}
This enables batching of observation vectors into fixed-dimension tensors for fast training, and upper bounds the evaluation time of $\fpi$ to guarantee real-time performance on hardware in large swarm experiments. We include an example observation encoding in Fig.~\ref{fig:overview}. 

\subsection{System Safety} 
We adopt the formulation of safe sets used in control barrier functions and define the global safe set $\cX$ as the super-level set of a global safety function $\fg: \cS \rightarrow \mathbb{R}$. We define this global safety as the minimum of local safety functions, $\fh: \cO \rightarrow \mathbb{R}$ that specify pairwise collision avoidance between all objects in the environment:
\begin{align}
    \cX &= \{ \vs \in \cS \ \ | \ \ \fg(\vs) > 0 \}, \label{eq:global_safety_set} \\ 
    \fg(\vs) &= \min_{i,j} \fh(\overline{\vp}\ji), \ \ \ 
    \fh(\overline{\vp}\ji) = \frac{ \| \overline{\vp}\ji \| - r_\mathrm{safe}}{ r_\mathrm{sense} - r_\mathrm{safe}},
\end{align}
where $\overline{\vp}\ji$ denotes the vector between the closest point on object $j$ to center of object $i$. This allows us to consistently define $r_\mathrm{safe}$ as the radius of the robot, where $r_\mathrm{safe} < r_\mathrm{sense}$. Intuitively, if a collision occurs between robots $i,j$, then $\fh(\overline{\vp}\ji)<0$ and $\fg(\vs)<0$,  implying that the system is not safe. In order to synthesize local controls with guaranteed global safety, we need to show non-local safety functions cannot violate global safety. Consider a pair of robots outside of the neighborhood, $\|\overline{\vp}\ji\| > r_\mathrm{sense}$. Clearly, $h(\overline{\vp}\ji) > 1 $, implying this interaction is always safe. 

\subsection{Controller Synthesis}
We use these safety functions to construct a global potential function, $\fpsi: \cS \rightarrow \mathbb{R}$ that becomes unbounded when any safety is violated, which resembles logarithmic barrier functions used in the interior point method in optimization~\cite{boyd_book}. Similarly, we can construct a local function, $\fpsi^i: \cO \rightarrow \mathbb{R}$: 
\begin{align}
	\fpsi(\vs) &= -\log \prod_i \prod_{j \in \cN^i} \fh(\overline{\vp}\ji), \\ 
	\fpsi^i(\vo^i) &= -\log \prod_{j \in \cN^i} \fh(\overline{\vp}\ji).
\end{align}
We use the local potential $\fpsi^i$ to synthesize the safety control module $\fb$.
We first state some assumptions. 
\begin{assumption}
    Initially, the distance between all objects is at least $r_\mathrm{safe} + \Delta_r$, where $\Delta_r$ is a user-specified parameter. \label{assumption: env reset}
\end{assumption}
\begin{assumption}
    \label{assumption:circle_robots}
    We assume robot $i$'s geometry not to exceed a ball of radius $r_\mathrm{safe}$ centered at $\vp^i$.
\end{assumption}

\begin{theorem}\label{theorem:si_safety}
For the single integrator dynamics~\eqref{eq:dynamics}, the safety defined by \eqref{eq:global_safety_set} is guaranteed under the control law~\eqref{eq:controller} with the following $ \fb(\vo^i)$ and $\fapi(\vo^i)$ for a scalar gains $k_p > 0$ and $k_c > 0$: 
\begin{align}
    \fb(\vo^i) &= -k_p \nabla_{\vp^i} \fpsi^i(\vo^i) \\
	\fapi(\vo^i) &= \begin{cases}
	    \frac{ (k_p - k_c) \|\nabla_{\vp^i}\fpsi^i(\vo^i)\|^2}{k_p\|\nabla_{\vp^i}\fpsi^i(\vo^i)\|^2 + |\langle \nabla_{\vp^i}\fpsi^i(\vo^i),\pi(\vo^i)\rangle|} & \Delta_h(\vo^i) < 0 \\ 
	    1 & \text{else}
	\end{cases} \label{eq:si_gain}  
\end{align} 
with $\Delta_h(\vo^i) = \min_{j\in\cN^i} \fh(\overline{\vp}\ji) - \Delta_r$, and $\nabla_{\vp^i}\fpsi^i$ as in~\eqref{eq:psigradp}.  	
\end{theorem}

\begin{IEEEproof}
We prove global safety by showing boundedness of $\fpsi^i(\vo^i)$, because a bounded $\fpsi(\vs)$ implies no safety violation. Since $\fpsi(\vs)$ is a sum of functions $\fpsi^i(\vo^i)$, it suffices to show that all $\fpsi^i(\vo^i)$ are bounded. To prove the boundedness of $\fpsi^i(\vo^i)$, we use a Lyapunov method. First, we note that $\fpsi^i(\vo^i)$ is an appropriate positive Lyapunov function candidate as $\fpsi^i(\vo^i) > 0 \ \forall \vo^i$, using the fact that $-\log(x) \in (0,\infty), \ \forall x \in(0,1)$. Second, we show $\vs \in \partial\cX^i \implies \dot{\fpsi}^i(\vo^i) < 0$, where the \emph{boundary-layer} domain $\partial\cX^i \subset \cS$ is defined as: 
\begin{align}
    \partial\cX^i = \{\vs \ | \ 0 < \min_{j\in\cN^i} \fh(\overline{\vp}\ji) < \Delta_r \}.
\end{align}
This result implies that upon entering the boundary-layer of the safe set, $\partial \cX^i$, the controller will push the system back into the interior of the safety set, $\cS \setminus \partial \cX^i$. 

We take the time derivative of $\fpsi^i$ along the system dynamics, and plug the single integrator dynamics~\eqref{eq:dynamics} and controller~\eqref{eq:controller} into $\dot{\fpsi}^i$: 
\begin{align}
    \dot{\fpsi^i} &= \langle \nabla_{\vs} \fpsi^i, \dot{\vs} \rangle
    = \sum_{i=1}^{n_i} \langle \nabla_{\vs^i} \fpsi^i, \dot{\vs}^i \rangle 
    = \sum_{i=1}^{n_i} \langle \nabla_{\vp^i} \fpsi^i, \vu^i \rangle \nonumber\\
    &= \sum_{i=1}^{n_i} \langle \nabla_{\vp^i} \fpsi^i, \fapi \pi + (1 - \fapi) \fb \rangle, \\ 
    &\text{ where } 
    \nabla_{\vp^i} \fpsi^i = \sum_{j\in\cN^i} \frac{ \overline{\vp}\ji}{ \|\overline{\vp}\ji\| (\|\overline{\vp}\ji\| - r_\mathrm{safe})}. \label{eq:psigradp}
\end{align}
From here, establishing that any element of the sum is negative when $\vs \in \partial\cX^i$ implies the desired result. Expanding an arbitrary element at the $i\th$ index with the definition of $\fb$:
\begin{align}
    \label{eq:si_element}
    &\langle \nabla_{\vp^i} \fpsi^i, \fapi \fpi + (1 - \fapi) \fb \rangle \\ 
    &= -k_p \| \nabla_{\vp^i} \fpsi^i \|^2 + \fapi ( \langle \nabla_{\vp^i} \fpsi^i, \fpi \rangle + k_p \| \nabla_{\vp^i} \fpsi^i \|^2) \nonumber\\ 
    &\leq -k_p \| \nabla_{\vp^i} \fpsi^i \|^2 + \fapi ( | \langle \nabla_{\vp^i} \fpsi^i, \fpi \rangle | + k_p \| \nabla_{\vp^i} \fpsi^i \|^2)\nonumber
\end{align}
By plugging $\fapi$ from~\eqref{eq:si_gain} into~\eqref{eq:si_element}, we arrive at the following: 
\begin{align}
    &\langle \nabla_{\vp^i} \fpsi^i, \fapi \fpi + (1 - \fapi) \fb \rangle \leq -k_c \| \nabla_{\vp^i} \fpsi^i \|^2 
\end{align}
Thus, every element is strictly negative, unless $\nabla_{\vp^i} \fpsi^i$ equals zero;  caused either by reaching the safety equilibrium of the system at $\fh(\overline{\vp}\ji)\geq 1, \forall j \in \cN^i$, or in the case of deadlock between robots.
\end{IEEEproof}

\begin{theorem}\label{theorem:di_safety}
For the double integrator dynamics given in \eqref{eq:dynamics}, the safety defined by \eqref{eq:global_safety_set} is guaranteed under control law \eqref{eq:controller} for the barrier-controller and the gain defined as: 
\begin{alignat}{2}
    \fb &= -k_v (\vv^i + k_p \nabla_{\vp^i} \fpsi^i) - k_p \frac{d}{dt} \nabla_{\vp^i} \fpsi^i - k_p \nabla_{\vp^i} \fpsi^i,\nonumber\\
	\fapi &= \begin{cases}
	    \frac{a_1 - k_c (k_p \fpsi^i + \frac{1}{2} \| \vv - \vk \|^2) }{a
	    _1 + |a_2|} \ \ \ \ & 
	    \Delta_h(\vo^i) < 0 \\
	    1 & \text{else}
	\end{cases}, \label{eq:di_gain} \\
	a_1 &= k_v \| \vv^i + k_p \nabla_{\vp^i} \fpsi^i \|^2 + k_p^2 \| \nabla_{\vp^i} \fpsi^i \|^2,\nonumber \\ 
	a_2 &= \langle \vv^i, k_p \nabla_{\vp^i} \fpsi^i \rangle + \langle \vv^i + k_p \nabla_{\vp^i} \fpsi^i, \fpi + k_p \frac{d}{dt} \nabla_{\vp^i} \fpsi^i \rangle, \nonumber
\end{alignat} 
where $k_p>0$, $k_c>0$, and $k_v>0$ are scalar gains, $\Delta_h$ is defined as in Theorem~\ref{theorem:si_safety}, $\frac{d}{dt} \nabla_{\vp^i} \fpsi^i$ is defined in~\eqref{eq:psigradp_dot} and the dependency on the observation is suppressed for legibility.
\end{theorem}

\begin{IEEEproof}
We take the same proof approach as in Theorem~\ref{theorem:si_safety} to show boundedness of $\psi^i$. We define a Lyapunov function, $\mathbb{V}$ augmented with a backstepping term: 
\begin{align}
    \mathbb{V} &= k_p \fpsi^i + \frac{1}{2} \| \vv - \vk \|^2,
\end{align}
where $\vv$ is the stacked velocity vector, $\vv = [\vv^1; \ldots ; \vv^{n_i}]$ and $\vk$ is the stacked nominally stabilizing control, $\vk = -k_p [ \nabla_{\vp^1} \fpsi^1;\ldots ; \nabla_{\vp^{n_i}} \fpsi^{n_i}]$. For the same reasoning as the previous result, $\mathbb{V}$ is a positive function, and thus an appropriate Lyapunov candidate. Taking the derivative along the system dynamics~\eqref{eq:dynamics}:
\begin{align}
    \dot{\mathbb{V}} 
    &= \sum_{i=1}^{n_i} k_p \langle \nabla_{\vs^i} \fpsi^i, \dot{\vs}^i \rangle + \langle \vv^i - \vk^i, \vu^i - \dot{\vk}^i \rangle \\
    &= \sum_{i=1}^{n_i} k_p \langle \nabla_{\vp^i} \fpsi^i, \vv^i \rangle + \langle \vv^i - \vk^i, \vu^i - \dot{\vk}^i \rangle.\label{eq:Vdot}
\end{align}
From here, establishing that an arbitrary element of the sum is negative when $\vs \in \partial\cX^i$ implies the desired result. We rewrite the first inner product in this expression as:
\begin{align}
    \langle \nabla_{\vp^i} \fpsi^i, \vv^i \rangle 
    &= \langle \nabla_{\vp^i} \fpsi^i, \vk^i \rangle + \langle \nabla_{\vp^i} \fpsi^i, (\vv^i - \vk^i) \rangle\nonumber \\
    &= -k_p \| \nabla_{\vp^i} \fpsi^i \|^2 + \langle \nabla_{\vp^i} \fpsi^i, (\vv^i - \vk^i) \rangle.
\end{align}
Next, we expand the second inner product of \eqref{eq:Vdot}, and plug $\vk^i$,$\vu^i$, and $\fb$ into~\eqref{eq:Vdot}:
\begin{align}
    &\langle \vv^i + k_p \nabla_{\vp^i} \fpsi^i, \vu^i + k_p \frac{d}{dt} \nabla_{\vp^i} \fpsi^i \rangle \\
    &= - k_v \| \vv^i + k_p \nabla_{\vp^i} \fpsi^i \|^2 - k_p \langle \nabla_{\vp^i} \psi^i, \vv^i + k_p \nabla_{\vp^i} \fpsi^i \rangle \nonumber\\ 
    & \hspace{2cm}  + \fapi \langle \vv^i + k_p \nabla_{\vp^i} \fpsi^i, \fpi-\fb \rangle \nonumber\\ 
    &\text{ where } \frac{d}{dt} \nabla_{\vp^i} \fpsi^i = 
    \sum_{j\in\mathcal{N}^i} \frac{\vv^i}{\|\overline{\vp}\ji\|(\|\overline{\vp}\ji\| - r_\mathrm{safe})} \nonumber \\ 
    -& \frac{\langle \vp^i, \vv^i \rangle \vp^i}{\|\overline{\vp}\ji\|(\|\overline{\vp}\ji\| - r_\mathrm{safe})(\|\overline{\vp}\ji\|^2 + (\|\overline{\vp}\ji\|-r_\mathrm{safe})^2)}
 \label{eq:psigradp_dot}
\end{align}
Combining both terms back into the expression: 
\begin{align}
    \label{eq:di_element}
    &k_p \langle \nabla_{\vp^i} \fpsi^i, \vv^i \rangle + \langle \vv^i - \vk^i, \vu^i - \dot{\vk}^i \rangle = -k_p^2 \| \nabla_{\vp^i} \fpsi^i \|^2 \nonumber\\ 
    -& k_v \| \vv^i + k_p \nabla_{\vp^i} \fpsi^i \|^2 + \fapi \langle \vv^i + k_p \nabla_{\vp^i} \fpsi^i, \fpi-\fb \rangle 
\end{align}
Expanding the last term with the definition of $\fb$, and upper bounding it: 
\begin{align}
    &\langle \vv^i + k_p \nabla_{\vp^i} \fpsi^i,\fpi-\fb \rangle \\ 
    \ \ &= (k_v \| \vv^i + k_p \nabla_{\vp^i} \fpsi^i \|^2 + k_p^2 \| \nabla_{\vp^i} \fpsi^i \|^2) \nonumber\\
    \ \ &+ (\langle \vv^i, k_p \nabla_{\vp^i} \fpsi^i \rangle + \langle \vv^i + k_p \nabla_{\vp^i} \fpsi^i, \fpi + k_p \frac{d}{dt} \nabla_{\vp^i} \fpsi^i \rangle)\nonumber \\
    \ \ &= a_1 + a_2 \leq a_1 + |a_2| \nonumber
\end{align}
where the terms in parentheses are grouped into $a_1,a_2$ to improve legibility. By plugging $\fapi$ from~\eqref{eq:di_gain} into~\eqref{eq:di_element}, we arrive at $
    \dot{\mathbb{V}} 
    = \sum_{i=1}^{n_i} -k_c \mathbb{V}$, which results in exponential stability that guarantees the system will remain safe by pushing it towards a safety equilibrium where $\fh(\overline{\vp}\ji)\geq 1, \forall j \in \cN^i$. It also makes the system robust to disturbances~\cite{Khalil_book}.
\end{IEEEproof}

We make some remarks on the results of the proofs. 
\begin{remark}
The setup for this proof is to give $\fpi$ maximal authority  without violating safety. This trade-off is characterized through the design parameter $\Delta_r$, and the gains $k_p$ and $k_v$ that control the measure of the conservativeness of the algorithm. For the discrete implementation of this algorithm, we introduce a parameter, $\varepsilon\ll 1$, to artificially decrease $\fapi$ such that $\fapi = 1 - \varepsilon$ when $\Delta_h > 0$. The results (in continuous time) of the proof still hold as $\fapi$ is a scalar gain on a destabilizing term, so a lower $\fapi$ further stabilizes the system. The gain $k_c$ can be arbitrarily small, in simulation we set it to $0$. 
\end{remark}
\begin{remark}
Intuitively, $\Delta_h(\vo^i)$ defines an unsafe domain for robot $i$. In safe settings, i.e. $\Delta_h > 0$, $\fapi=1-\varepsilon$ and so the barrier has little effect on the behavior. In most unsafe cases, the barrier will be activated, driving a large magnitude safety response. However, in dense multi-robot settings, it is possible for safety responses to cancel each other out in a gridlock, resulting in dangerous scenarios where small disturbances can cause the system to violate safety. In this case, we use the above result to put an adaptive gain, $\fapi$~(\ref{eq:si_gain},\ref{eq:di_gain}) on the neural policy and to drive $\fapi$ to $0$, cancelling the effect of $\fpi$. Thus, we use a convex combination of the neural optimal policy and the safety control module to guarantee safety in all cases. 
\end{remark}
\begin{remark}
In Theorem~\ref{theorem:si_safety}, we synthesize a local nominal control that guarantees the global safety of the system. In Theorem~\ref{theorem:di_safety}, we use Lyapunov backstepping to provide the same nominal control through a layer of dynamics. This method is valid for a large class of nonlinear dynamical systems known as full-state feedback linearizeable systems~\cite{Khalil_book}. By following this method, GLAS can be extended to other nonlinear dynamical systems in a straightforward manner. 
\end{remark}

\begin{figure*}[t]
    \centering
    \begin{subfigure}{0.19\textwidth}
        \centering
        \includegraphics[width=\textwidth]{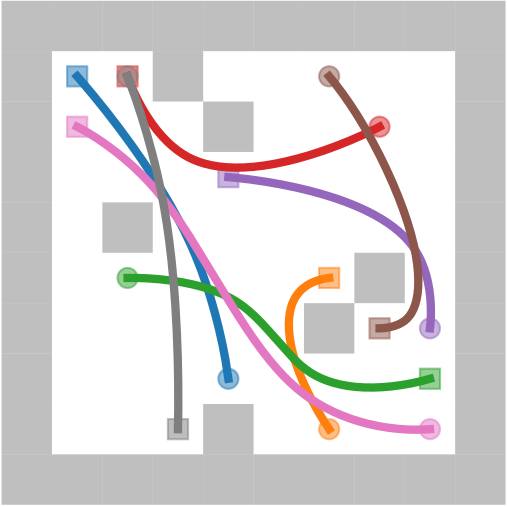}
        \subcaption{Global}
        \label{fig:ss_ex_global}
    \end{subfigure}%
    \hfill
    \begin{subfigure}{0.19\textwidth}
        \centering
        \includegraphics[width=\textwidth]{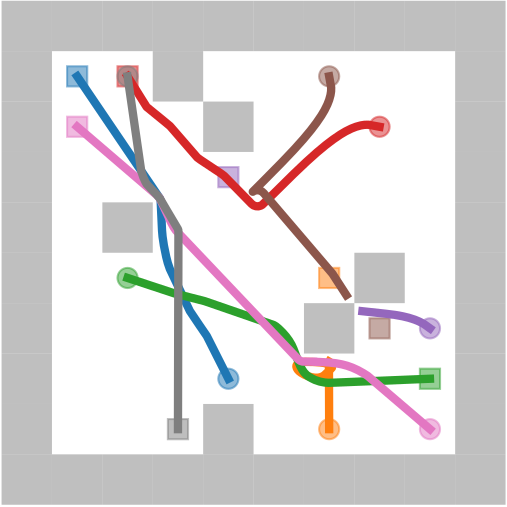}
        \subcaption{ORCA}
        \label{fig:ss_ex_orca}
    \end{subfigure}%
    \hfill
    \begin{subfigure}{0.19\textwidth}
        \centering
        \includegraphics[width=\textwidth]{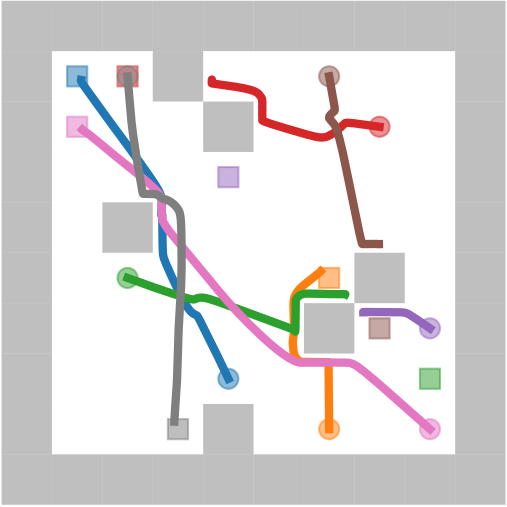}
        \subcaption{GLAS Barrier}
        \label{fig:ss_ex_barrier}
    \end{subfigure}%
    \hfill
    \begin{subfigure}{0.19\textwidth}
        \centering
        \includegraphics[width=\textwidth]{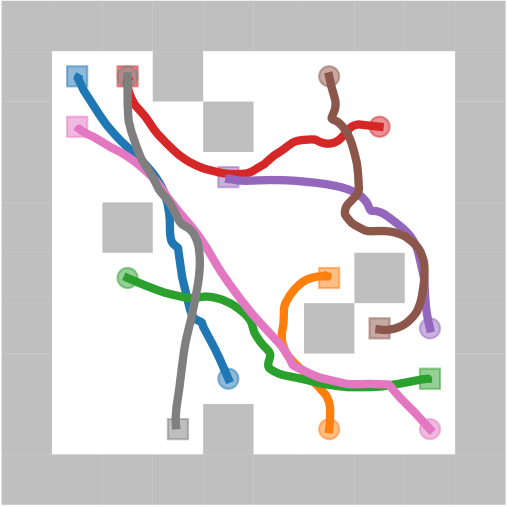}
        \subcaption{GLAS Two-stage}
        \label{fig:ss_ex_twostage}
    \end{subfigure}%
    \hfill
    \begin{subfigure}{0.19\textwidth}
        \centering
        \includegraphics[width=\textwidth]{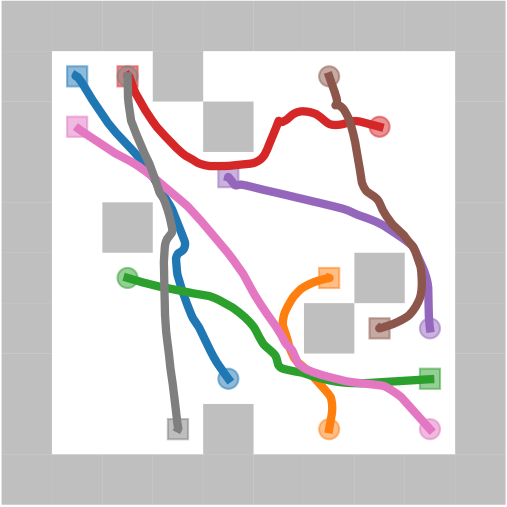}
        \subcaption{GLAS End-to-end}
        \label{fig:ss_ex_endtoend}
    \end{subfigure}
    \caption{Example trajectories for baselines (a-c) and our proposed method (d,e), where the goal is to move robots from their starting position (circles) to the goal position (squares). Our methods achieve the highest success rate. The GLAS end-to-end policy generates trajectories that use less control effort.}
    \label{fig:ss_ex}
\end{figure*}

\section{Experiments}
\label{sec:experiments}
We now present results of simulation comparing GLAS and its variants with state-of-the-art baselines as well as experimental results on physical quadrotors. Our supplemental video includes additional simulations and experiments.

\subsection{Learning Implementation and Hyperparameters}
For data generation, we use an existing implementation of a centralized global trajectory planner~\cite{Hoenig_2018} and generate $\approx2$$\times10^5$ (\SI{200}{\thousand}) demonstrations in random \SI{8x8}{m} environments with 10 or \SI{20}{\percent} obstacles randomly placed in a grid pattern and 4, 8, or 16 robots (e.g., see Fig.~\ref{fig:ss_ex} for \SI{10}{\percent} obstacles and 8 robots). We sample trajectories every \SI{0.5}{s} and generate $|\cD|=40$$\times 10^6 \ (\SI{40}{\million})$ data points in total, evenly distributed over the 6 different environment kinds. We use different datasets for single and double integrator dynamics with different desired smoothness in the global planner. 

We implement our learning framework in Python using PyTorch~\cite{pyTorch}. The $\fphi_\cOmega$ and $\fphi_\cV$ networks have an input layer with 2 neurons, one hidden layer with 64 neurons, and an output layer with 16 neurons. The $\frho_\cOmega$ and $\frho_\cV$ networks have 16 neurons in their input and output layers and one hidden layer with 64 neurons. The $\fPsi$ network has an input layer with 34 neurons, one hidden layer with 64 neurons, and outputs $\fpi$ using two neurons. All networks use a fully connected feedforward structure with ReLU activation functions. We use an initial learning rate of 0.001 with the PyTorch optimizer ReduceLROnPlateau function, a batch size of \SI{32}{\thousand}, and train for 200 epochs. During manual, iterative hyperparameter tuning, we found that the hidden layers should at least use 32 neurons. For efficient training of the Deep Set architecture, we create batches where the number of neighbors $|\cN_\cV|$ and number of obstacles $|\cN_\cOmega|$ are the same and limit the observation to a maximum of 6 neighbors and 6 obstacles.

\subsection{GLAS Variants}

We study the effect of each component of the system architecture by comparing variants of our controller: \emph{end-to-end}, \emph{two-stage}, and \emph{barrier}. 
End-to-end and two-stage are synthesized through~\eqref{eq:controller}, but differ in how $\fpi$ is trained. 
For end-to-end we calculate the loss on $\fu(\vo^i)$, while for two-stage we calculate the loss on $\fpi(\vo^i)$. Comparing these two methods isolates the effect of the end-to-end training. 
The barrier variant is a linear feedback to goal controller with our safety module. Essentially, barrier is synthesized with~\eqref{eq:controller}, where the $\fpi$ heuristic is replaced with a linear goal term: $K \ve^{ii}$, where, for single integrator systems, $K = k_p I$, and for double integrator systems, $K = [k_p I, k_v I]$, with scalar gains $k_p$ and $k_v$. Studying the performance of the barrier variant isolates the effect of the global-to-local heuristic training.

\subsection{Single Integrator Dynamics}
\label{subsubsection:si_eval}

We compare our method with ORCA, a state-of-the-art decentralized approach for single integrator dynamics. Unlike GLAS, ORCA requires relative velocities with respect to neighbors in addition to relative positions. All methods compute a velocity action with guaranteed safety. 

We show example trajectories for the global planner, ORCA, and GLAS variants in Fig.~\ref{fig:ss_ex}. In Fig.~\ref{fig:ss_ex_orca}/\ref{fig:ss_ex_barrier}, the purple and brown robots are getting stuck in local minima caused by obstacle traps. In Fig.~\ref{fig:ss_ex_twostage}/\ref{fig:ss_ex_endtoend}, our learned policies are able to avoid those local minima. 
The end-to-end approach produces smoother trajectories that use less control effort, e.g., red and brown robot trajectories in Fig.~\ref{fig:ss_ex_endtoend}. 

\subsubsection{Evaluation of Metrics}
\label{subsubsection:si_eval_metrics}
\begin{figure}
    \centering
    \includegraphics[width=\linewidth]{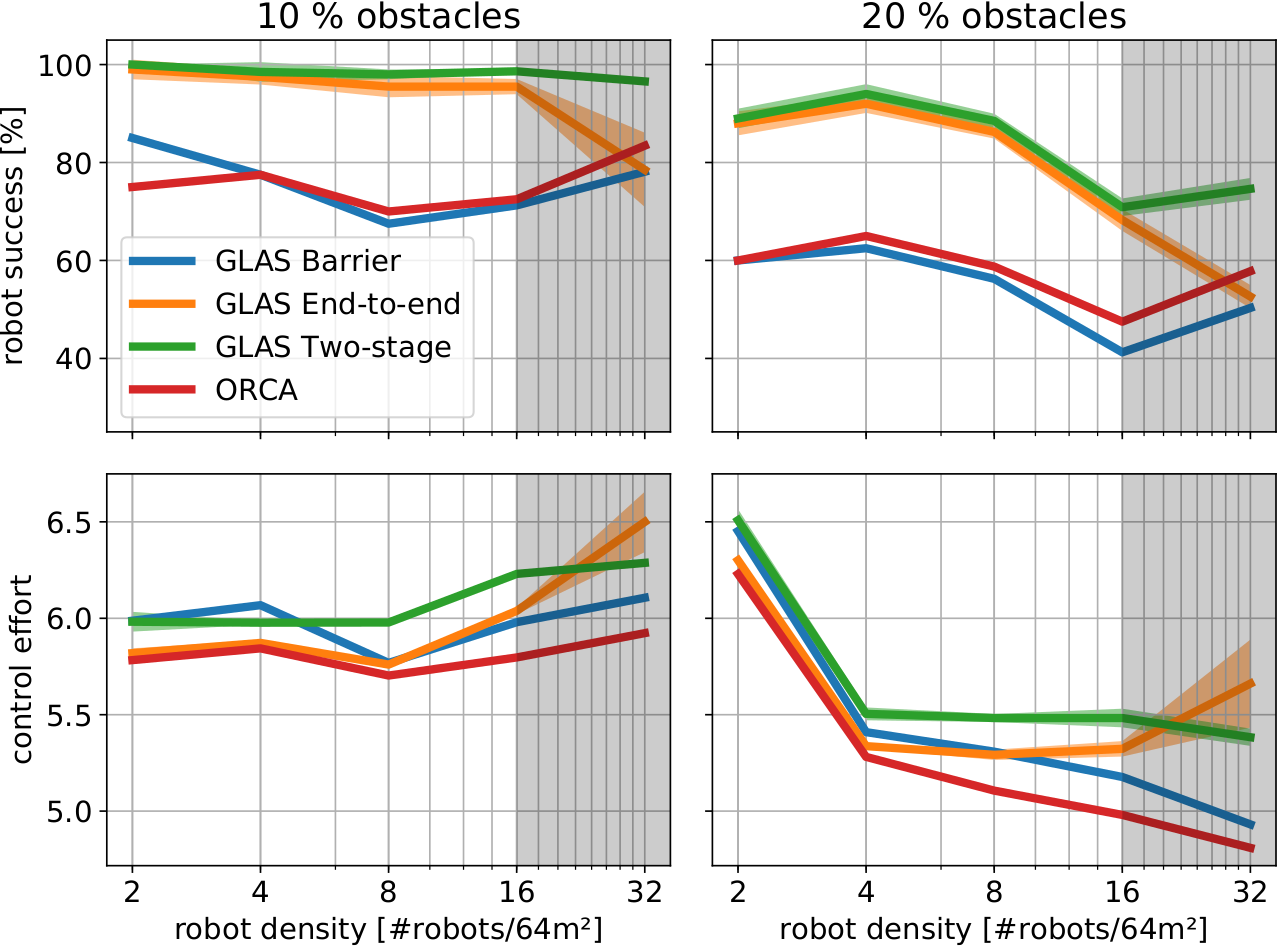}
    \caption{Success rate and control effort with varying numbers of robots in a \SI{8 x 8}{m} space for single integrator systems. Shaded area around the lines denotes standard deviation over 5 repetitions. The shaded gray box highlights validation outside the training domain.}
    \vspace{-5mm}
    \label{fig:exp1}
\end{figure}

We deploy the baseline and variants over 100 validation cases with 2, 4, 8, 16, and 32 robots and \SI{10}{\percent} and \SI{20}{\percent} obstacle density (10 validation instances for each case) and empirically evaluate the metrics defined in \eqref{eq:metrics}.
Our training data only contains different examples with up to 16 robots. We train 5 instances of end-to-end and two-stage models, to quantify the effect of random-weight initialization in the neural networks on performance, see Fig.~\ref{fig:exp1}.

In the top row, we consider the success metric $r_s$. In a wide range of robot/obstacle cases (2--16 robots/\SI{64}{m^2}), our global-to-local methods outperform ORCA by \SI{20}{\percent}, solving almost all instances. Our barrier variant has a similar success rate as ORCA, demonstrating that the neural heuristic $\pi$ is crucial for our high success rates.
The two-stage approach generalizes better to higher-density cases beyond those in the training data. We observe the inverse trend in the double integrator case and analyzing this effect is an interesting future direction.

In the bottom row, we measure control effort  $r_p$. Our end-to-end approach uses less control effort than the two-stage approach. ORCA has the lowest control effort, because the analytical solution to single integrator optimal control is a bang-bang controller, similar in nature to ORCA's implementation. 

\subsubsection{Effect of Complexity of Data on Loss Function}
\begin{figure}
    \centering
    \includegraphics[width=\linewidth]{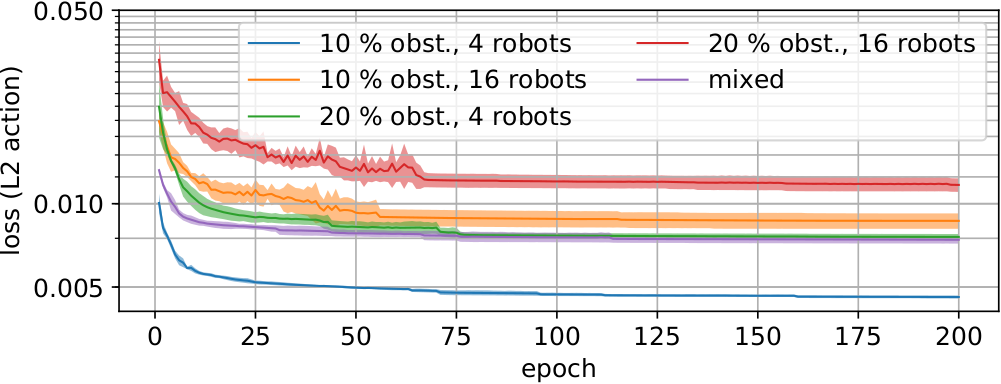}
    \caption{Testing loss when training using 10 and \SI{20}{\percent} obstacles and 4 or 16 robots. Synthesizing a distributed policy that is consistent with the global data is harder for high robot densities than for high obstacle densities. We use the GLAS end-to-end with $|\cD|=\SI{5}{\million}$ and repeat 5 times.}
    \label{fig:exp3loss}
\end{figure}
At some robot/obstacle density, local observations and their actions will become inconsistent, i.e., the same observation will match to different actions generated by the global planner. We quantify this data complexity limit by recording the value of the validation loss when training using datasets of varying complexity, see Fig.~\ref{fig:exp3loss}. 

Here, we train the end-to-end model with $|\cD|=\SI{5}{\million}$ using isolated datasets of 10 and \SI{20}{\percent} obstacles and with 4 and 16 robots, as well as a mixed dataset. We see that the easiest case, 4 robots and \SI{10}{\percent} obstacles, results in a the smallest loss, roughly \SI{1}{\percent} of the maximum action magnitude of the expert. The learning task is more difficult with high robot density compared to high obstacle density. A mixed dataset, as used in all other experiments, is a good trade-off between imitating the expert very well and being exposed to complex situations.

\subsubsection{Effect of Radius of Sensing on Performance}
\begin{figure}
    \centering
    \includegraphics[width=\linewidth]{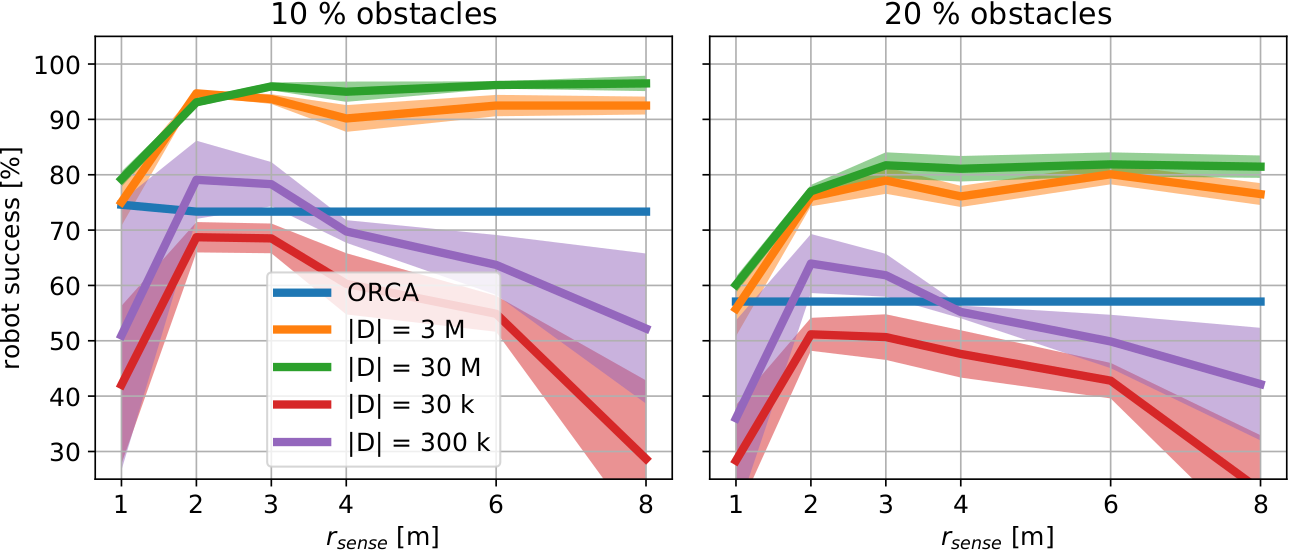}
    \caption{Effect of sensing radius and amount of training data on robot success rate. The validation has 4, 8, and 16 robot cases with 10 instances each. Training and validation were repeated 5 times; the shaded area denotes the standard deviation.}
    \label{fig:global_to_local_learning}
\end{figure}
We quantify the transition from local-to-global by evaluating the performance of models trained with various sensing radii and dataset size. We evaluate performance on a validation set of 4, 8, and 16 robot cases with \SI{10}{\percent} and \SI{20}{\percent} obstacle densities.
First, we found that there exists an optimal sensing radius for a given amount of data, which increases with larger datasets. For example, in the \SI{20}{\percent} obstacle case, the optimal sensing radius for $|\cD|=\SI{300}{\thousand}$ is around \SI{2}{m} and the optimal radius for $|\cD|=\SI{30}{\million}$ is \SI{8}{m}. Second, we found that between models of various dataset sizes the performance gap at small sensing radii is smaller compared to the performance gap at large sensing radii. This result suggests that little data is needed to use local information well, and large amounts of data is needed to learn from global data. 

\subsection{Double Integrator Dynamics}
\begin{figure}
    \centering
    \includegraphics[width=\linewidth]{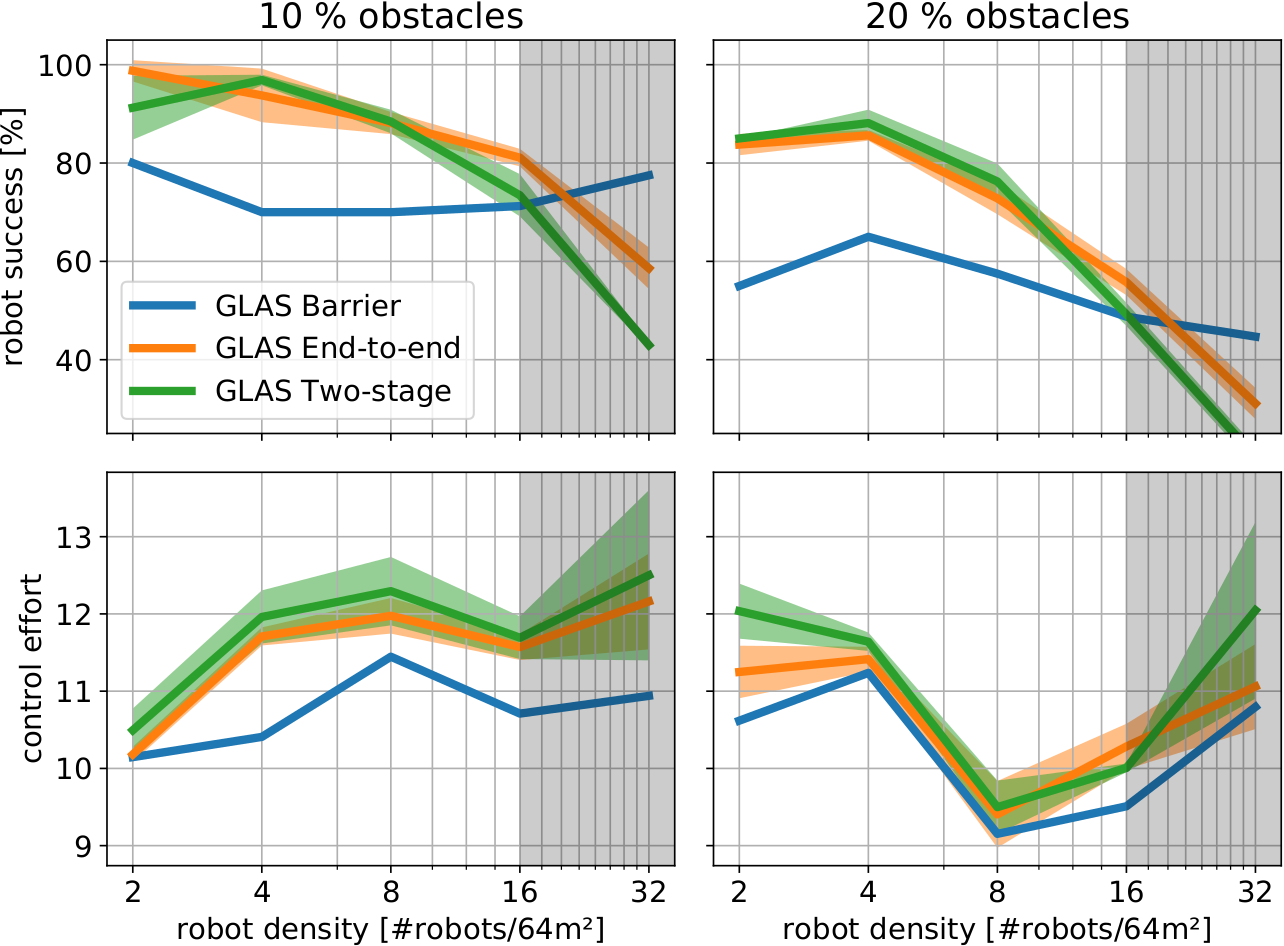}
    \caption{Success rate and control effort with varying numbers of robots in a \SI{8 x 8}{m} workspace for double integrator systems. Shaded area around the lines denotes standard deviation over 5 repetitions. The shaded gray box highlights validation outside the training domain.}
    \vspace{-5mm}
    \label{fig:exp1_di}
\end{figure}

We extend our results to double integrator dynamical systems to demonstrate GLAS extends naturally as a dynamically-coupled motion planner. Similar to the single integrator evaluation, we show double integrator statistical evaluation with respect to the performance metrics \eqref{eq:metrics}, for varying robot and obstacle density cases. We use the same setup as in the single integrator case (see Sec.~\ref{subsubsection:si_eval}), but with a different dataset ($|\cD|=\SI{20}{\million}$). Results are shown in Fig.~\ref{fig:exp1_di}. 

In the top row, we consider the success metric $r_s$. In a wide range of robot density cases (2--16 robots/\SI{64}{m^2}), our global-to-local end-to-end method again outperforms the barrier baseline by \SIrange{15}{20}{\percent}. In the 32 robots/\SI{64}{m^2} case, the barrier baseline outperforms the global-to-local methods. We suspect this is a combination of our method suffering from the significantly higher complexity of the problem and the naive barrier method performing well because the disturbances from the robot interaction push the robot out of local minima obstacle traps. In contrast to the single integrator case, the end-to-end solution generalizes better than the two-stage in higher robot densities. We conjecture that having a much larger training set can significantly improve performance.

In the bottom row, we consider the performance metric $r_p$. For double integrator systems, the cost function $\fc(\vs,\vu)$ corresponds to the energy consumption of each successfully deployed robot. The end-to-end variant uses less effort ($\leq \SI{6.25}{\percent}$) than the two-stage method on average. 

\subsection{Experimental Validation with Aerial Swarms}
We implement the policy evaluation ($\fpi,\fapi,\fb$) in C to enable real-time execution on-board of Crazyflie 2.0 quadrotors using double integrator dynamics (see Fig.~\ref{fig:overview}).
The quadrotors use a small STM32 microcontroller with \SI{192}{kB} SRAM running at \SI{168}{MHz}.
Our policy evaluation takes \SI{3.4}{ms} for 1 neighbor and \SI{5.0}{ms} for 3 neighbors, making it computationally efficient enough to execute our policy in real-time at \SI{40}{Hz}.
On-board, we evaluate the policy, forward-propagate double integrator dynamics, and track the resulting position and velocity setpoint using a nonlinear controller. The experimental validation demonstrates that our policy generalizes to novel environments where the obstacles are arranged in continuous space, as opposed to on a grid. 

We use a double integrator GLAS end-to-end policy in three different scenarios with up to 3 obstacles and 12 quadrotors. We fly in a motion capture space, where each robot is equipped with a single marker, using the Crazyswarm~\cite{crazyswarm} for tracking and scripting.
Our demonstration shows that our policy works well on robots and that it can also handle cases that are considered difficult in decentralized multi-robot motion planning, such as swapping positions with a narrow corridor.

\section{Conclusion} 
In this work, we present GLAS, a novel approach for multi-robot motion planning that combines the advantages of existing centralized and distributed approaches. Unlike traditional distributed methods, GLAS avoids local minima in many cases. Unlike existing centralized methods, GLAS only requires local relative state observations, which can be measured on board or communicated locally. 
We propose an end-to-end training approach using a novel differentiable safety method compatible with general dynamical models, resulting in a dynamically-coupled motion planner with guaranteed collision-free operation and empirically-validated low control effort solutions. 
In future work, we will explore adaptive data-set aggregation methods~\cite{ross2011reduction} for more efficient expert querying. We will also compare the computational effort and performance of the Deep Set representation with deep CNN methods as well as Graph Neural Network methods~\cite{scarselli2008graph}.

\printbibliography

\end{document}